\documentclass[conference]{IEEEtran}
\IEEEoverridecommandlockouts

\usepackage{cite}
\usepackage{amsmath,amssymb,amsfonts}
\usepackage{algorithmic}
\usepackage{graphicx}
\usepackage{textcomp}
\usepackage{xcolor}
\usepackage{booktabs}
\usepackage{multirow}
\usepackage{array}
\usepackage{hyperref}

\def\BibTeX{{\rm B\kern-.05em{\sc i\kern-.025em b}\kern-.08em
    T\kern-.1667em\lower.7ex\hbox{E}\kern-.125emX}}

\begin{document}

\title{NeR-SC: Adapting Neural Video Representation to Screen Content}

\author{\IEEEauthorblockN{Ruohan Shi}
\IEEEauthorblockA{\textit{Management school} \\
\textit{The University of Sheffield}\\
srhandlrj@gmail.com}
\and
\IEEEauthorblockN{Jiaoyan Zhao}
\IEEEauthorblockA{\textit{Undergraduate School of Artificial Intelligence} \\
\textit{Shenzhen Polytechnic University}\\
zhaojy@szpu.edu.cn}
\and
\IEEEauthorblockN{Haogang Feng*}
\IEEEauthorblockA{\textit{School of Artificial Intelligence} \\
\textit{Shenzhen University of Information Technology}\\
fenghaogang@suit-sz.edu.cn}
\thanks{*Corresponding author.}
}

% \author{\IEEEauthorblockN{Qingyu Mao}
% \IEEEauthorblockA{\textit{School of Electronic and Information Engineering} \\
% \textit{Shenzhen University}\\
% qingyu.mao@outlook.com}
% % \and
% % \IEEEauthorblockN{Youneng Bao}
% % \IEEEauthorblockA{\textit{School of Electronic and Information Engineering} \\
% % \textit{Shenzhen University}\\
% % baoyn@szu.edu.cn}
% \and
% \IEEEauthorblockN{Yongsheng Liang*}
% \IEEEauthorblockA{\textit{School of Electronic and Information Engineering} \\
% \textit{Shenzhen University}\\
% liangys@szu.edu.cn}
% \thanks{*Corresponding author.}
% }

\maketitle

\begin{abstract}
Implicit neural representations have emerged as a promising paradigm for video compression, with recent methods achieving competitive performance on natural video.
However, screen content video---common in remote desktop, online education, and cloud gaming---exhibits distinct statistics: sharp edges, limited color palettes, and strong temporal redundancy.
Existing neural representation methods, designed for natural scenes, lack mechanisms to exploit these properties, leaving substantial room for improvement.
In this paper, we propose NeR-SC, a neural representation framework tailored for screen content video. Building on the SNeRV backbone, NeR-SC introduces three screen-content-specific modules: (i) a learnable color palette that models the discrete color structure of screen content by restricting the low-frequency sub-band to a learned color set; (ii) a multi-gate dense fusion module that replaces sequential feature fusion with dense, attention-gated cross-stage interaction; and (iii) an embedding-level frame skip strategy that bypasses redundant decoder invocations for static frames, with zero training overhead.
Experiments on DSCVC and VCD show that NeR-SC achieves 40.32~dB and 41.73~dB average PSNR, outperforming representative neural video representation methods and, at low bitrates, surpassing H.264 and H.265. The skip strategy enables real-time decoding with no loss in quality.
\end{abstract}


\section{Introduction}

Screen content video---pervasive in remote desktop, online education, and cloud gaming---differs fundamentally from natural video. It exhibits sharp edges at text and UI elements, large uniform regions with limited color palettes, and strong temporal redundancy where most pixels remain static across extended frame sequences~\cite{hevc_scc_overview}.

Traditional coding standards have addressed these properties through specialized tools such as Intra Block Copy and palette coding (HEVC-SCC, VVC, AV1~\cite{scc_standards}). Deep learning-based screen content codecs (DSCVC~\cite{dscvc}, DSCIC~\cite{dscic}) confirm the value of content-adaptive design, but rely on conventional prediction--residual architectures with constrained decoding throughput.

In parallel, neural video representation overfits a small network to a single video so that the network weights become the compressed representation~\cite{nerv}. Subsequent works (HNeRV~\cite{hnerv}, HiNeRV~\cite{hinerv}, FFNeRV~\cite{ffnerv}, DS-NeRV~\cite{dsnerv}, SNeRV~\cite{snerv}) have steadily advanced quality on natural benchmarks. However, these methods lack three capabilities critical for screen content: (1) they regress continuous RGB without exploiting limited color palettes; (2) their sequential feature fusion is suboptimal for multi-scale graphical content; and (3) they decode every frame independently, ignoring temporal redundancy.

In this paper, we propose \textbf{NeR-SC} (Neural Representation for Screen Content), a framework that extends the SNeRV backbone~\cite{snerv} with three screen-content-specific innovations. Our main contributions are:

\begin{enumerate}
    \item We propose a \textbf{learnable color palette} that restricts the low-frequency sub-band to a discrete set of learned colors, converting continuous regression into per-pixel classification while preserving full expressiveness for high-frequency details.

    \item We introduce a \textbf{multi-gate dense fusion (MGF)} module that replaces sequential pair-wise fusion with dense, simultaneous fusion across decoder stages, augmented by Squeeze-and-Excitation residual gating.

    \item We design an \textbf{embedding-level frame skip} strategy that detects temporal redundancy via compact latent embeddings, safely bypassing the decoder for static frames with zero additional training cost.

    \item Experiments on DSCVC~\cite{dscvc} and VCD~\cite{vcd} show that NeR-SC achieves 40.32~dB and 41.73~dB, outperforming representative neural video representation methods and, at low bitrates, surpassing H.264 and H.265. The skip strategy enables real-time decoding with no loss in quality.
\end{enumerate}

\begin{figure*}[t]
    \centering
    \includegraphics[width=\textwidth]{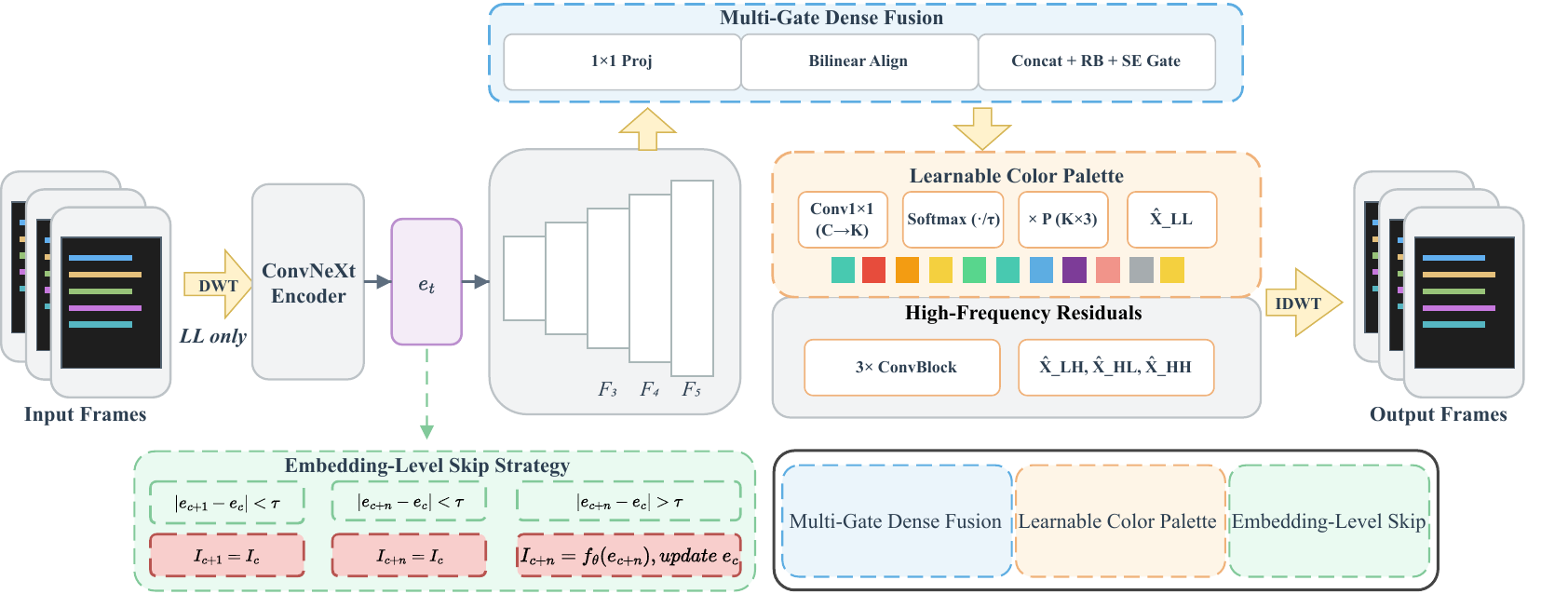}
    \caption{Overview of the proposed NeR-SC framework.}
    \label{fig:architecture}
\end{figure*}

\section{Related Work}

\subsection{Neural Video Representation}

Neural video representation overfits a small neural network to a single video so that the network weights become the compressed representation~\cite{nerv,inric}. NeRV~\cite{nerv} pioneered this paradigm for video, and subsequent works have steadily advanced quality and efficiency through content-adaptive embeddings (HNeRV~\cite{hnerv}), hierarchical positional encodings (HiNeRV~\cite{hinerv}), and improved temporal modeling (FFNeRV~\cite{ffnerv}, DS-NeRV~\cite{dsnerv}, T-NeRV~\cite{tnerv}). SNeRV~\cite{snerv} introduced wavelet-based frequency decomposition. Extensions to other domains (NeRV360~\cite{nerv360}, MV-HiNeRV~\cite{mvhinerv}) and efficiency-focused refinements (E-NeRV~\cite{enerv}, OnlineReP~\cite{onlinerepnerv}, LT-NeRV~\cite{ltnerv}) have further broadened the scope of neural representation. Despite these advances, existing methods are designed for natural scenes and lack specialized mechanisms for the sharp edges, limited color palettes, and strong temporal redundancy unique to screen content.

\subsection{Screen Content Video Coding}

The distinct statistics of screen content render conventional transform-based codecs inefficient, as the DCT energy compaction assumption breaks for non-continuous tone regions~\cite{hevc_scc_overview}. HEVC-SCC addressed this with Intra Block Copy, palette coding, and adaptive color transform, and similar tools have been adopted in VVC and AV1~\cite{scc_standards,av1_overview,vvc_scc}. Deep learning-based screen content codecs such as DSCVC~\cite{dscvc} and DSCIC~\cite{dscic} further confirm the value of content-adaptive design, but remain grounded in hybrid prediction--residual architectures. Our work addresses this gap by adapting neural video representation to the statistical properties of screen content.

\section{Method}

In this section, we present NeR-SC, a neural representation framework designed for screen content video. We first briefly describe the overall architecture built upon the SNeRV backbone~\cite{snerv}, then introduce three key innovations: a learnable color palette, a multi-gate dense fusion module, and an embedding-level frame skip strategy.

\subsection{Overall Architecture}

The overall architecture of NeR-SC is illustrated in Fig.~\ref{fig:architecture}. Our framework adopts the SNeRV backbone~\cite{snerv}: an input frame $X$ is decomposed via a single-level 2D Haar wavelet transform,
\begin{equation}
X_{LL}, X_{LH}, X_{HL}, X_{HH} = \text{Haar}(X),
\label{eq:haar}
\end{equation}
the LL sub-band is encoded through a ConvNeXt encoder~\cite{liu2022convnet} into a latent embedding $\mathbf{z}_i$ and decoded through PixelShuffle upsampling blocks~\cite{shi2016real}, and the three HF sub-bands are reconstructed via lightweight convolutional heads. The full frame is recovered by inverse transform:
\begin{equation}
\hat{X} = \text{IWT}\big(\hat{X}_{LL}, [\hat{X}_{LH}, \hat{X}_{HL}, \hat{X}_{HH}]\big).
\label{eq:hfr}
\end{equation}
This design processes only the LL sub-band ($\frac{1}{4}$ resolution) through the heavy encoder-decoder, while the HFR heads add high-frequency details as lightweight post-processing.

Building upon this backbone, we introduce three innovations: (i) a learnable color palette (Section~\ref{sec:palette}) that replaces unconstrained RGB regression with a discrete palette for the LL sub-band; (ii) a multi-gate dense fusion module (Section~\ref{sec:mgf}) that replaces sequential pair-wise fusion with dense, attention-gated cross-stage interaction; and (iii) an embedding-level skip strategy (Section~\ref{sec:skip}) that bypasses redundant decoder invocations for static frames with zero training overhead.

\subsection{Learnable Color Palette}
\label{sec:palette}

Screen content uses a limited number of distinct colors (e.g., tens of syntax-highlighting colors in a code editor), in sharp contrast to the continuous color distributions of natural images. This discrete-color property is concentrated in the LL sub-band (smooth regions, uniform fills), whereas the HF sub-bands encode transitions and gradients that require continuous values. We therefore apply a learnable color palette exclusively to the LL sub-band, keeping the HFR heads unconstrained to preserve edge fidelity.

We introduce a trainable palette matrix $P \in \mathbb{R}^{K \times 3}$ ($K$=64). Given the fused decoder features $F_{\text{dec}}$, a $1{\times}1$ convolution produces $K$-channel logits $L$, followed by a temperature-scaled softmax:
\begin{equation}
A_{k,i,j} = \frac{\exp(L_{k,i,j} / \tau)}{\sum_{m=1}^{K} \exp(L_{m,i,j} / \tau)}.
\label{eq:palette_softmax}
\end{equation}
The LL sub-band is reconstructed as a convex combination of palette entries:
\begin{equation}
\hat{X}_{LL}(i,j) = \sum_{k=1}^{K} A_{k,i,j} \cdot \mathbf{p}_k,
\label{eq:palette_einsum}
\end{equation}
then denormalized and combined with HFR outputs via inverse DWT (Eq.~\ref{eq:hfr}). By converting continuous regression into per-pixel classification where screen content is most color-sparse, the palette yields an average PSNR improvement of 0.22 dB in our ablation study (Table~\ref{tab:ablation}).

\subsection{Multi-Gate Dense Fusion}
\label{sec:mgf}

The SNeRV decoder fuses multi-scale features through a sequential pair-wise MRF strategy~\cite{snerv}, which restricts information flow to adjacent stages and relies on fixed transposed-convolution upsamplers. We propose the Multi-Gate Dense Fusion (MGF) module, which replaces this sequential cascade with dense, simultaneous fusion of the last three decoder stages $(F_3, F_4, F_5)$. Per-stage $1{\times}1$ convolutions project all features to a common dimension $C_{\text{fuse}}$, coarser features are aligned via bilinear interpolation, and the concatenated features are fused through a residual block:
\begin{equation}
F_{\text{fuse}} = \text{RB}\big([\text{Up}(F'_3); \text{Up}(F'_4); \text{Up}(F'_5)]).
\label{eq:mgf_concat}
\end{equation}
A Squeeze-and-Excitation (SE) block~\cite{senet} applies per-channel attention in a residual manner:
\begin{equation}
F_{\text{out}} = F_{\text{fuse}} + F_{\text{fuse}} \odot \sigma(\text{FC}(\text{ReLU}(\text{FC}(\text{GAP}(F_{\text{fuse}}))))).
\label{eq:mgf_se}
\end{equation}
Compared to MRF, MGF enables direct cross-stage information flow, provides lighter alignment via bilinear interpolation, and learns per-channel importance weights through SE residual gating. In our ablation (Table~\ref{tab:ablation}), MGF adds 0.17 dB PSNR gain while reducing parameters (1.52M $\rightarrow$ 1.49M).

\subsection{Embedding-Level Skip Strategy}
\label{sec:skip}

Screen content videos exhibit strong temporal redundancy: changes across frames are typically confined to small regions (e.g., cursor movement), while the majority of pixels remain static. Since the decoding function $\mathcal{D}(\cdot)$ is deterministic, frames with similar latent embeddings $\mathbf{z}_i$ yield near-identical reconstructions. Our insight is that compact embeddings ($16 \times 2 \times 4$ elements for $960 \times 1920$) can serve as efficient temporal redundancy proxies---comparing them is orders of magnitude cheaper than full decoding.

During decoding, we compare $\mathbf{z}_i$ to the embedding $\mathbf{z}_k$ of the last decoded frame. If the normalized $L_1$ distance falls below threshold $\tau$, the decoder is bypassed:
\begin{equation}
\hat{I}_i =
\begin{cases}
\hat{I}_k, & \text{if } \frac{1}{|\mathbf{z}|}\|\mathbf{z}_i - \mathbf{z}_k\|_1 < \tau, \\[4pt]
\mathcal{D}(\mathbf{z}_i; \theta_{\mathcal{D}}), & \text{otherwise}.
\end{cases}
\end{equation}
The reference $\mathbf{z}_k$ is updated only on full decodes, preventing cumulative drift. The mechanism introduces zero training overhead, stores only a single scalar $\tau$ (0.005), and boosts decoding throughput by 20.7 FPS with no PSNR loss, enabling real-time decoding (61.7 FPS).

\section{Experiments}
\label{sec:experiments}

\subsection{Datasets}
We evaluate on two screen-content datasets. \textbf{VCD}~\cite{vcd} contains video-conferencing clips in three categories: talking head (TH), talking head with opaque background (THOB), and talking head with background blur (THBB); we select two sequences per category (6 videos, 1,800 frames). \textbf{DSCVC}~\cite{dscvc} comprises diverse screen-content sequences; we use ten sequences (video\_11--video\_20, 2,102 frames). All videos are center-cropped to 960$\times$1920~\cite{hnerv}.

\subsection{Implementation Details}

NeR-SC is implemented in PyTorch and trained on an NVIDIA RTX 4090 GPU. The model uses the SNeRV backbone with five encoder-decoder stages, Haar wavelet decomposition, a default parameter budget of $\sim$3.0M, palette size $K{=}64$, and MGF fusing the last three decoder stages. We use Adam ($lr{=}1{\times}10^{-3}$), cosine annealing, batch size of 8, 300 epochs, and L2 loss. The skip threshold $\tau$ is $0.005$.

\subsection{Compared Methods and Evaluation Metrics}

We compare NeR-SC against NeRV~\cite{nerv}, HNeRV~\cite{hnerv}, and SNeRV~\cite{snerv}, all re-implemented from official code and re-trained on our datasets (learning rates 0.0008 for NeRV/HNeRV; $+1$e$-$8 bias in SNeRV wavelet computation). Conventional codecs H.264/x264~\cite{x264} and H.265/x265~\cite{x265} (\texttt{medium} preset) serve as reference baselines. Quality is assessed via PSNR and MS-SSIM, storage via bits per pixel (bpp), and RD performance via PSNR-vs-bpp curves.

\section{Results and Discussion}

\subsection{Quantitative Results}
Table~\ref{tab:results} reports PSNR (dB) and MS-SSIM on Video~16 under varying model sizes (1.5M--6.0M) and training epochs (300--1800). NeR-SC consistently achieves the best results across all configurations. In the model-size study, the margin over SNeRV reaches 1.41~dB at 6.0M, and NeR-SC at 1.5M (38.54~dB) already exceeds HNeRV at 6.0M (36.81~dB). In the epoch study, NeR-SC maintains a lead at every stage, confirming that the gain stems from architectural design rather than longer optimization.

\begin{table}[t]
\centering
\caption{Quantitative results on `Video\_16' under (a) different model sizes and (b) different training epochs.}
\label{tab:results}

\vspace{1mm}
{\small (a) Different model sizes} \\
\vspace{1mm}
\resizebox{\linewidth}{!}{
\begin{tabular}{l|cccc|c}
\toprule
 Method & 1.5M & 3.0M & 4.5M & 6.0M & Avg. \\
\midrule
NeRV~\cite{nerv}   & 28.08/.9052 & 30.21/.9380 & 31.91/.9558 & 33.62/.9686 & 30.96/.9419 \\
HNeRV~\cite{hnerv} & 30.46/.9381 & 34.16/.9681 & 35.51/.9752 & 36.81/.9803 & 34.24/.9654 \\
SNeRV~\cite{snerv} & 38.11/.9880 & 40.91/.9927 & 41.79/.9940 & 41.92/.9940 & 40.68/.9922 \\ \midrule
NeR-SC & \textbf{38.54/.9888} & \textbf{41.12/.9929} & \textbf{42.37/.9944} & \textbf{43.33/.9956} & \textbf{41.34/.9929} \\
\bottomrule
\end{tabular}}

\vspace{3mm}
{\small (b) Different training epochs} \\
\vspace{1mm}
\resizebox{\linewidth}{!}{
\begin{tabular}{l|cccc|c}
\toprule
 Method & 300 & 600 & 1200 & 1800 & Avg. \\
\midrule
NeRV~\cite{nerv}   & 30.21/.9380 & 33.52/.9680 & 35.79/.9786 & 36.47/.9811 & 34.00/.9664 \\
HNeRV~\cite{hnerv} & 34.16/.9681 & 38.76/.9852 & 40.74/.9893 & 41.43/.9930 & 38.77/.9839 \\
SNeRV~\cite{snerv} & 40.91/.9927 & 42.04/.9944 & 42.73/.9950 & 42.96/.9958 & 42.16/.9946 \\
NeR-SC & \textbf{41.12/.9929} & \textbf{42.17/.9945} & \textbf{42.84/.9955} & \textbf{43.13/.9958} & \textbf{42.32/.9947} \\
\bottomrule
\end{tabular}}
\end{table}

Table~\ref{tab:full} presents per-sequence results across all 16 test videos. NeR-SC achieves the best average PSNR on both datasets: 40.32 dB (DSCVC) and 41.73 dB (VCD), outperforming SNeRV by 0.54 dB and 0.11 dB. Gains are particularly notable on Video\_16 (41.07 vs. 40.91 dB) and Video\_17 (41.93 vs. 41.10 dB), where the palette and MGF modules benefit from rich structural and color regularities.

On VCD, NeR-SC trails SNeRV only on THOB sequences (40.40 vs. 40.47 dB; 43.22 vs. 43.29 dB). THOB features a human speaker overlaid on a screen background, where natural face statistics (continuous skin tones, soft shading) are less aligned with the palette's discrete-color prior. The gap is marginal ($\leq$0.07 dB), and NeR-SC retains the highest VCD average, demonstrating robustness across diverse content.

\begin{table}[t]
    \centering
    \caption{Per-sequence PSNR (dB)/MS-SSIM results on DSCVC and VCD datasets.}
    \label{tab:full}
    \resizebox{\linewidth}{!}{
    \small
    \begin{tabular}{l|ccc|c}
    \toprule
    Video & NeRV~\cite{nerv} & HNeRV~\cite{hnerv} & SNeRV~\cite{snerv} & NeR-SC(Ours) \\
    \midrule
    Video\_11 & 26.16/.9875 & 35.92/.9986 & 38.34/.9992 & \textbf{38.58/.9993} \\
    Video\_12 & 23.47/.9595 & 34.67/.9944 & 36.60/.9967 & \textbf{36.87/.9967} \\
    Video\_13 & 21.06/.9041 & 29.46/.9682 & 31.22/.9735 & \textbf{31.49/.9755} \\
    Video\_14 & 26.67/.9729 & 36.62/.9954 & 38.67/.9972 & \textbf{38.89/.9975} \\
    Video\_15 & 33.17/.9775 & 38.14/.9907 & 39.60/.9956 & \textbf{39.90/.9959} \\
    Video\_16 & 30.19/.9378 & 34.24/.9686 & 40.91/.9927 & \textbf{41.07/.9928} \\
    Video\_17 & 31.51/.9424 & 35.93/.9702 & 41.10/.9917 & \textbf{41.93/.9928} \\
    Video\_18 & 29.57/.9313 & 31.72/.9401 & 37.25/.9918 & \textbf{37.79/.9929} \\
    Video\_19 & 39.17/.9945 & 46.72/.9985 & 49.59/.9994 & \textbf{49.79/.9995} \\
    Video\_20 & 41.76/.9936 & 46.23/.9966 & 44.47/.9984 & \textbf{46.87/.9988} \\
    \midrule
    DSCVC Avg. & 30.27/.9601 & 36.97/.9821 & 39.78/.9936 & \textbf{40.32/.9942} \\
    \midrule
    th001   & 32.27/.9751 & 34.96/.9872 & 35.76/.9918 & \textbf{35.78/.9920} \\
    th002   & 36.31/.9634 & 37.98/.9739 & 39.34/.9820 & \textbf{39.38/.9821} \\
    thbb001 & 41.04/.9808 & 42.71/.9865 & 43.61/.9906 & \textbf{44.34/.9922} \\
    thbb002 & 43.87/.9902 & 45.79/.9932 & 47.23/.9960 & \textbf{47.25/.9960} \\
    thob001 & 35.32/.9734 & 38.48/.9849 & \textbf{40.47/.9929} & 40.40/.9927 \\
    thob002 & 36.01/.9801 & 41.06/.9925 & \textbf{43.29/.9961} & 43.22/.9961 \\
    \midrule
    VCD Avg. & 37.47/.9772 & 40.16/.9864 & 41.62/.9916 & \textbf{41.73/.9919} \\
    \bottomrule
    \end{tabular}}
    \end{table}

\begin{figure}[t]
    \centering
    \includegraphics[width=0.9\linewidth]{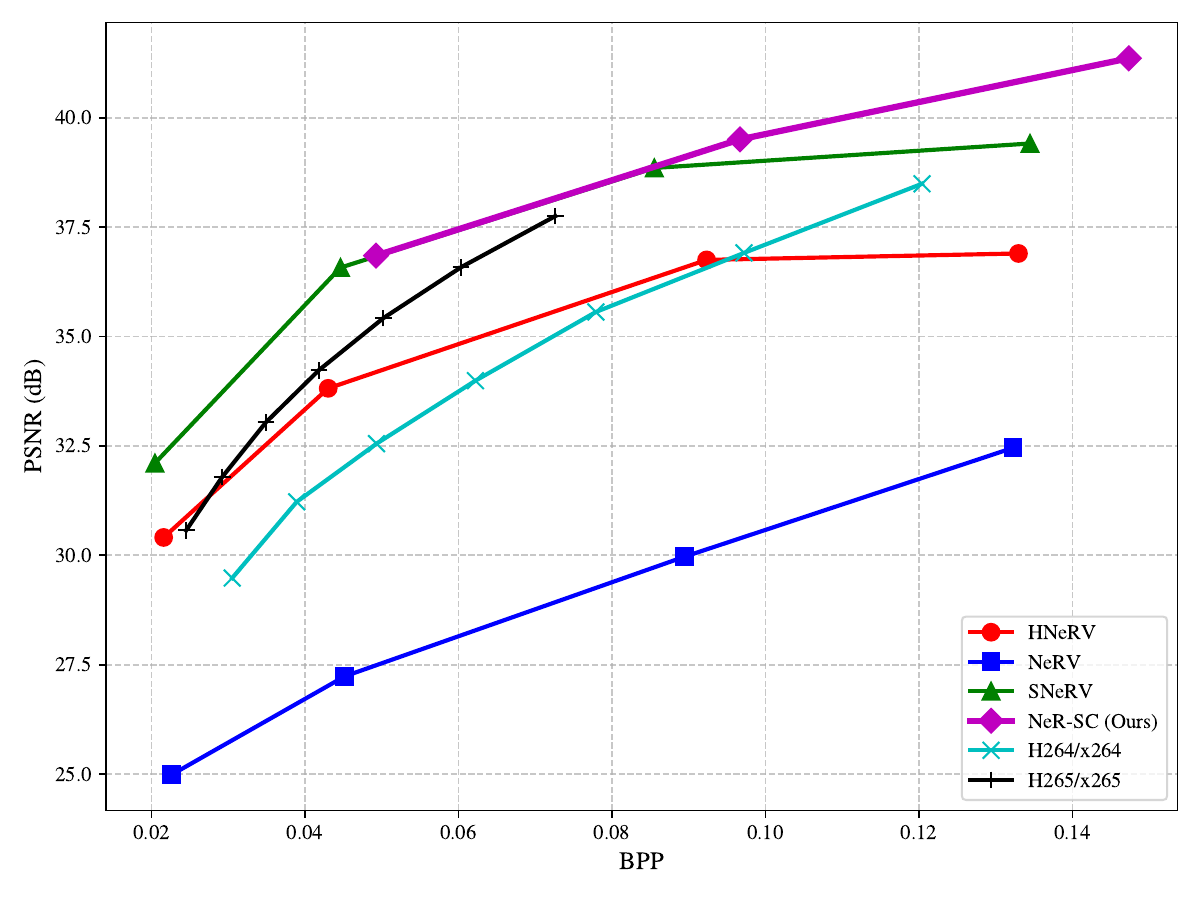} \\
    \vspace{0.5em}
    \includegraphics[width=0.9\linewidth]{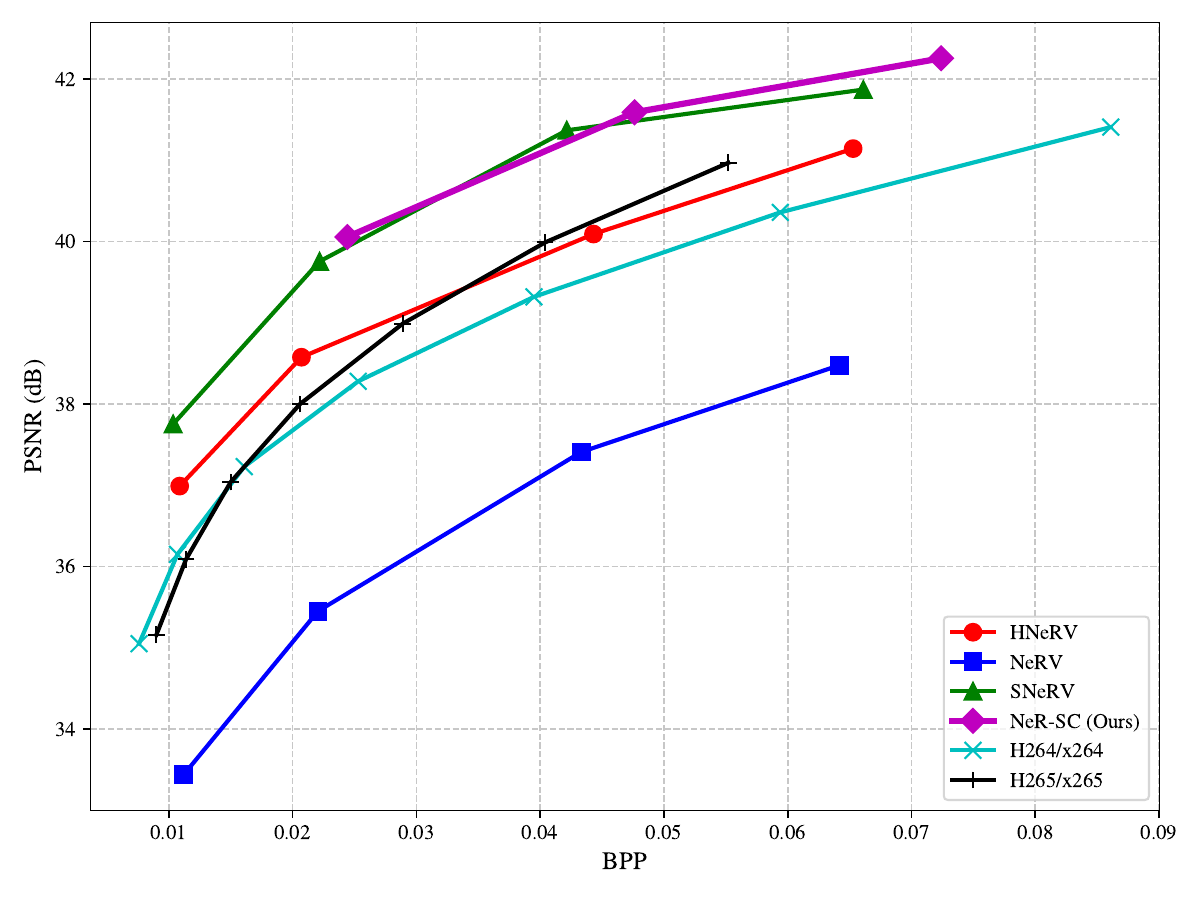}
    \caption{Rate-distortion curves (PSNR vs.\ bpp) on the DSCVC (top) and VCD (bottom) datasets.}
    \label{fig:rd_curves}
\end{figure}

Fig.~\ref{fig:rd_curves} presents the rate-distortion (RD) curves (PSNR vs.\ bpp) on the DSCVC and VCD datasets. NeR-SC consistently outperforms all other neural video representation methods across the entire bitrate range, with a substantial margin over HNeRV and NeRV. Notably, at low bitrates on both datasets, NeR-SC surpasses H.264 and H.265, demonstrating that the proposed palette, MGF, and skip modules collectively enable highly efficient compression tailored to screen content statistics.

\subsection{Qualitative Results}

\begin{figure*}[t]
    \centering
    \includegraphics[width=\textwidth]{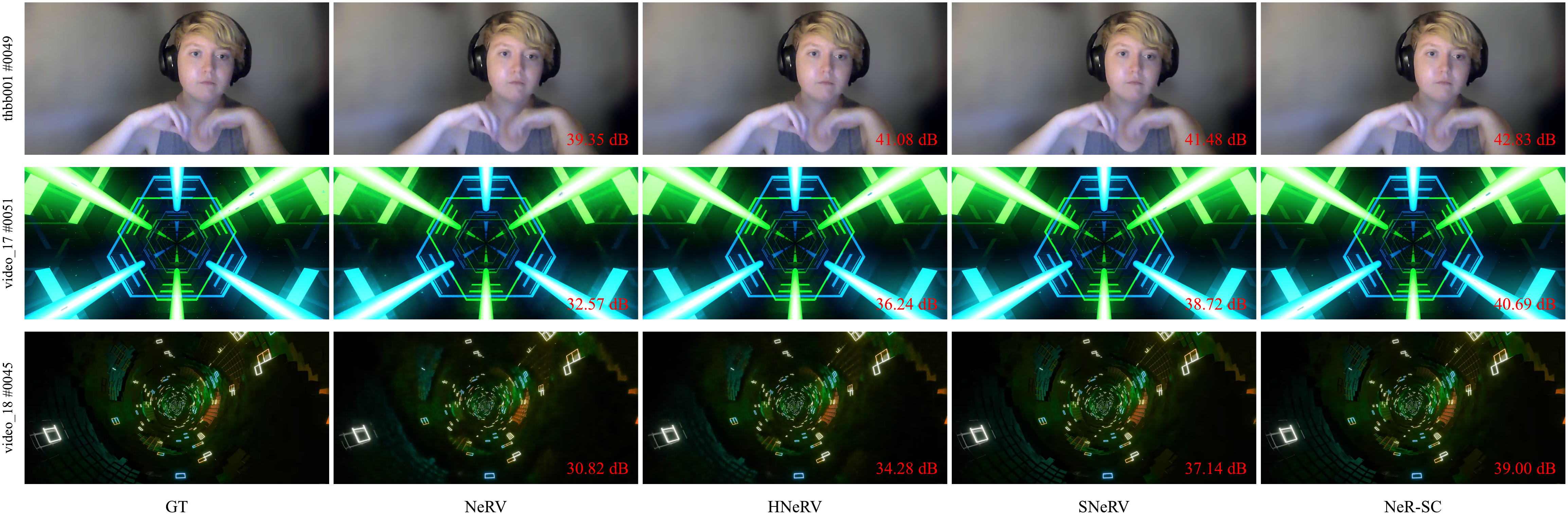}
    \caption{Qualitative comparison of reconstructed frames across four methods and ground truth on three cases.}
    \label{fig:qualitative}
\end{figure*}

Fig.~\ref{fig:qualitative} shows visual comparisons across four methods and the ground truth on three representative cases from the DSCVC dataset. NeRV and HNeRV exhibit clear artifacts and blurred edges, particularly around high-contrast boundaries, reflecting their lack of specialization for screen content. Both SNeRV and NeR-SC recover sharper structures thanks to their wavelet-based frequency decomposition. However, in the second row (Video\_17), SNeRV omits several content elements---notably the dark rectangular blocks in the top-right and bottom-left regions---while NeR-SC faithfully reconstructs them with well-preserved boundaries. Across all cases, NeR-SC produces the cleanest rendering and most consistent color reproduction, attributable to the palette module's discrete color prior and the MGF module's enhanced multi-scale feature fusion.

\subsection{Ablation Studies}

\subsubsection{Component Ablation}

To validate the contribution of each proposed component, we conduct an incremental ablation study on three representative screen content videos. Table~\ref{tab:ablation} summarizes the reconstruction quality (PSNR) and decoding throughput (FPS) for each configuration.

Starting from the baseline, adding the Palette module improves average PSNR by 0.22 dB, with a 4.9 FPS drop from the color table lookup. Adding MGF contributes a further 0.17 dB (0.39 dB total) while recovering 0.8 FPS, as narrower channel widths reduce parameters (1.52M $\rightarrow$ 1.49M). The skip strategy adds 20.7 FPS with no PSNR loss, lifting NeR-SC to 61.7 FPS for real-time decoding with zero training overhead.

\begin{table}[htbp]
\centering
\caption{Ablation study of NeR-SC components on DSCVC dataset (PSNR$\uparrow$/FPS$\uparrow$).}
\label{tab:ablation}
\resizebox{\linewidth}{!}{
\begin{tabular}{l|ccc|cc}
\toprule
Config. & video\_16 & video\_17 & video\_18 & Avg. & $\Delta$ Avg.\\
\midrule
Baseline  & 38.14/39.9 & 38.42/41.7 & 33.48/41.5 & 36.68/41.0 & ---/---\\
+ Palette & 38.28/36.0 & 38.56/35.7 & 33.85/36.8 & 36.90/36.1 & +0.22/-4.9\\
+ MGF     & 38.37/41.6 & 38.97/41.8 & 33.88/41.9 & 37.07/41.8 & +0.39/+0.8\\
+ Skip    & \textbf{38.37/56.1} & \textbf{38.97/71.6} & \textbf{33.88/57.4} & \textbf{37.07/61.7} & \textbf{+0.39/+20.7}\\
\bottomrule
\end{tabular}}
\end{table}

\subsubsection{Palette Size Ablation}

We further investigate the impact of palette size $K$ on compression performance. Table~\ref{tab:palette_ablation} reports PSNR (dB) and MS-SSIM on three representative screen content videos under varying $K$ from 16 to 128. All models are trained at 1.5M parameters for 100 epochs to isolate the effect of palette capacity.

\begin{table}[htbp]
\centering
\caption{Palette size ablation on selected screen content videos. All models are trained at 1.5M parameters for 100 epochs.}
\label{tab:palette_ablation}
\begin{tabular}{l|c|c|c|c}
\toprule
Video & $k{=}16$ & $k{=}32$ & $k{=}64$ & $k{=}128$ \\
%  & PSNR & MS-SSIM & PSNR & MS-SSIM & PSNR & MS-SSIM & PSNR & MS-SSIM & PSNR & MS-SSIM \\
\midrule
video\_16 & 32.84/0.973 & 32.49/0.971 & \textbf{33.11/0.975} & 32.65/0.973 \\
video\_17 & \textbf{33.18/0.973} & 32.89/0.973 & 32.14/0.971 & 32.25/0.971 \\
video\_18 & 22.36/0.881 & 23.54/0.906 & \textbf{23.78/0.909} & 22.89/0.884 \\
\midrule
Avg. & 29.46/0.942 & 29.64/0.950 & \textbf{29.68/0.952} & 29.26/0.943 \\
\bottomrule
\end{tabular}
\end{table}

Average PSNR and MS-SSIM improve consistently as $K$ increases from 16 to 64 (29.46 dB to 29.68 dB), confirming that larger palette capacity benefits screen content. The gain is content-dependent: video\_18 with richer UI elements benefits most ($+1.42$ dB), while simpler videos show marginal improvements. Performance degrades at $K{=}128$ (29.26 dB)---under a fixed 1.5M parameter budget, the expanded embedding dimension reduces decoder capacity and introduces optimization difficulty. We adopt $K{=}64$ for all experiments.

\subsubsection{Embedding-Level Frame Skip}

\begin{figure}[t]
    \centering
    \begin{minipage}{0.8\linewidth}
        \includegraphics[width=\linewidth]{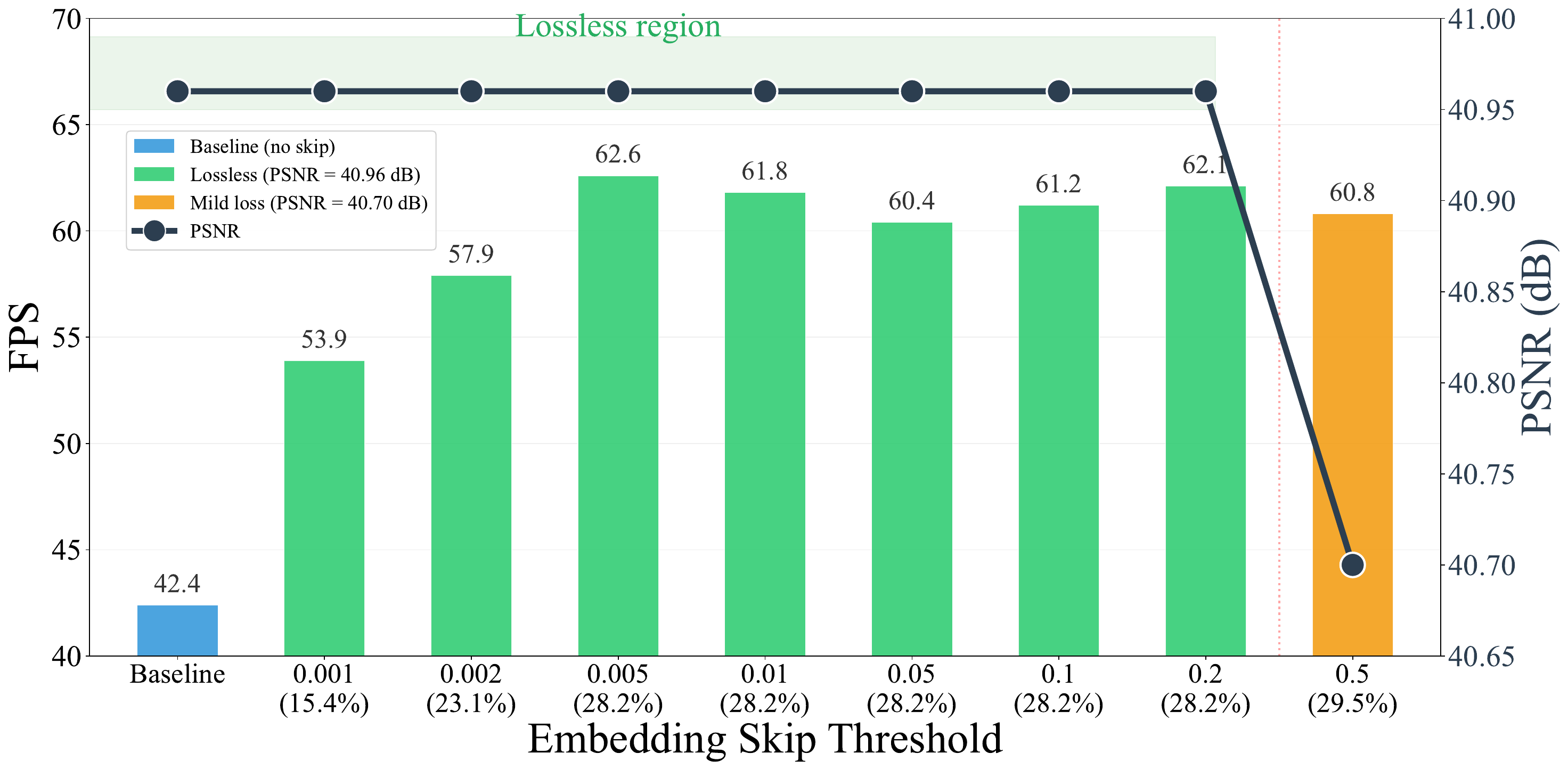}
        \centerline{(a) video\_16}
    \end{minipage}
    
    \vspace{1em}
    \begin{minipage}{0.8\linewidth}
        \includegraphics[width=\linewidth]{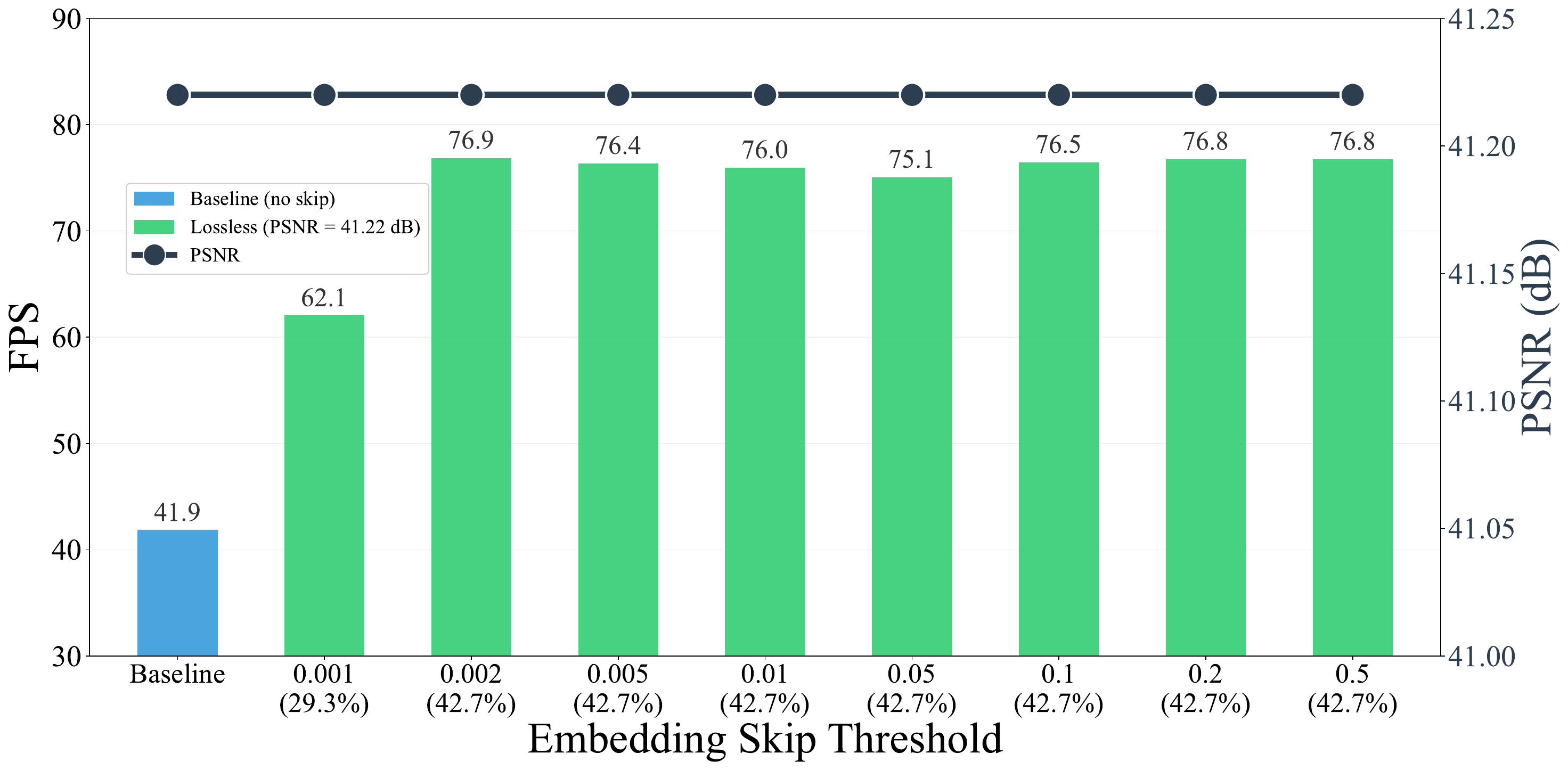}
        \centerline{(b) video\_17}
    \end{minipage}
    \caption{Trade-off between decoding throughput and reconstruction quality under varying embedding skip thresholds on (a) video\_16 and (b) video\_17. Values in parentheses along the x-axis denote the frame skip rate at each threshold.}
    \label{fig:skip_tradeoff}
\end{figure}

We further evaluate the embedding skip strategy on video~16 and video~17 using a pre-trained SNeRV (3.0M). Fig.~\ref{fig:skip_tradeoff} shows the FPS--PSNR trade-off across skip thresholds $\tau$, with frame skip rates annotated along the x-axis.

In the lossless region ($\tau \leq 0.05$), the skip strategy raises throughput from 42.4 to 62.6 FPS on video\_16 (+47.6\%) and from 62.1 to 76.9 FPS on video\_17 with no measurable quality loss. The skip rate saturates at $\sim$28\% and $\sim$43\%, respectively, representing each video's inherent temporal redundancy bound. Beyond $\tau=0.05$, video\_16 enters a mild quality-loss regime while video\_17 remains lossless. We set $\tau=0.005$ as the default, which can be tuned per-video for latency-sensitive or quality-critical applications.

\section{Conclusion}

In this paper, we introduced NeR-SC, a neural representation framework specifically designed for screen content video. By building upon the SNeRV backbone and incorporating three screen-content-specific modules---a learnable color palette, multi-gate dense fusion, and an embedding-level skip strategy---NeR-SC effectively addresses the unique challenges of screen content: limited color palettes, the need for multi-scale feature interaction, and strong temporal redundancy. Experimental results on DSCVC and VCD datasets demonstrate that NeR-SC achieves 40.32~dB and 41.73~dB average PSNR, outperforming SNeRV by 0.54~dB and 0.11~dB. At low bitrates on both datasets, NeR-SC surpasses H.264 and H.265, confirming that neural representation methods, when properly adapted to the target content type, can compete with---and even outperform---traditional codecs for screen content. The skip strategy further enables real-time decoding at 61.7~FPS with no loss in quality. Future work will explore extending NeR-SC to higher resolutions and integrating with traditional SCC tools for hybrid compression pipelines.

\section*{Acknowledgment}
This work was supported by the Shenzhen Polytechnic University Research Fund (No. 6025310035K).

\bibliographystyle{IEEEtran}
\bibliography{references}

\end{document}